\documentclass{vgtc}                          

\graphicspath{{figures/}{pictures/}{images/}{./}} 

\usepackage{times}                     

\usepackage{tabu}                      
\usepackage{booktabs}                  
\usepackage{lipsum}                    
\usepackage{mwe}                       
\usepackage{listings}
\usepackage{color}  

\usepackage{caption} 
\newcommand\PB{\textit{PowerBoost}}

\definecolor{codepink}{rgb}{0,0.49411764705882355,0.7764705882352941}
\definecolor{codegray}{rgb}{0.5,0.5,0.5}
\definecolor{codepurple}{rgb}{0.58,0,0.82}
\definecolor{backcolour}{rgb}{1,0.9764705882352941,0.984313725490196}

\lstdefinestyle{mystyle}{
    backgroundcolor=\color{backcolour},   
    commentstyle=\color{codepink},
    keywordstyle=\color{magenta},
    numberstyle=\tiny\color{codegray},
    stringstyle=\color{codepurple},
    basicstyle=\footnotesize,
    breakatwhitespace=false,         
    breaklines=true,                 
    captionpos=b,                    
    keepspaces=true,                 
    numbers=left,                    
    numbersep=5pt,                  
    showspaces=false,                
    showstringspaces=false,
    showtabs=false,                  
    tabsize=2
}

\lstdefinelanguage{JavaScript}{
  keywords={break, case, catch, continue, debugger, default, delete, do, else, finally, for, if, in, instanceof, new, return, switch, this, throw, try, typeof, var, void, while, with},
  morekeywords={Boostlet, init, select_box, load_script, send_http_post, get_image, set_image, set_mask, convert_to_png, filter, hint, select_seed},
  morecomment=[l]{//},
  morecomment=[s]{/*}{*/},
  morestring=[b]',
  morestring=[b]",
  ndkeywords={class, export, boolean, throw, implements, import, this},
  keywordstyle=\color{codepink}\bfseries,
  ndkeywordstyle=\color{darkgray}\bfseries,
  identifierstyle=\color{black},
  stringstyle=\color{codepurple}\ttfamily,
  sensitive=true
}

\usepackage{mathptmx}                  
\usepackage{fontawesome5}

\onlineid{0}

\vgtccategory{Research}

\vgtcinsertpkg

\title{\faBookmark \hspace{1mm} Boostlet.js: Image processing plugins for the web\\ via JavaScript injection}

\author{Edward Gaibor\thanks{e-mail: edward.gaibor001@umb.edu}\\ %
        \scriptsize University of Massachusetts Boston %
\and Shruti Varade\thanks{e-mail: s.varade001@umb.edu}\\ %
     \scriptsize University of Massachusetts Boston %
\and Rohini Deshmukh\thanks{e-mail: r.deshmukh001@umb.edu}\\ %
     \scriptsize University of Massachusetts Boston %
\and Tim Meyer\thanks{e-mail: tim.meyer@unibw.de}\\ %
     \scriptsize University of the Bundeswehr Munich %
\and Mahsa Geshvadi\thanks{e-mail: mahsa.geshvadi001@umb.edu}\\ %
     \scriptsize University of Massachusetts Boston %
\and SangHyuk Kim\thanks{e-mail: sanghyuk.kim001@umb.edu}\\ %
     \scriptsize University of Massachusetts Boston %
\and Vidhya Sree Narayanappa\thanks{e-mail: v.narayanappa001@umb.edu}\\ %
     \scriptsize University of Massachusetts Boston %
\and Daniel Haehn\thanks{e-mail: daniel.haehn@umb.edu}\\ %
     {\scriptsize University of Massachusetts Boston}}

\teaser{
  \centering
  \includegraphics[width=\linewidth]{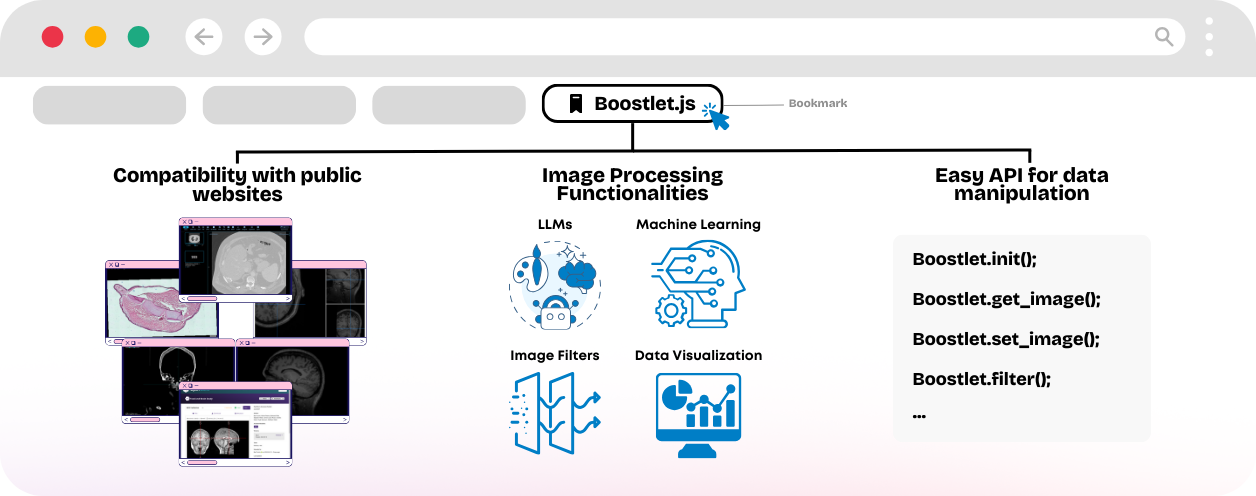}
  \caption{The Boostlet.js library provides a suite of image processing plugins for the web that integrates within various available visualization frameworks using JavaScript injection. This library enables direct access to image data via a simple and easy-to-use API and can be deployed as bookmarklets. Examples, code, and data are available at \href{https://boostlet.org}{https://boostlet.org}.}
  \label{fig:teaser}
}

\abstract{
    Can web-based image processing and visualization tools easily integrate into existing websites without significant time and effort? Our Boostlet.js library addresses this challenge by providing an open-source, JavaScript-based web framework to enable additional image processing functionalities. Boostlet examples include kernel filtering, image captioning, data visualization, segmentation, and web-optimized machine-learning models. To achieve this, Boostlet.js uses a browser bookmark to inject a user-friendly plugin selection tool called \PB~into any host website. Boostlet also provides on-site access to a standard API independent of any visualization framework for pixel data and scene manipulation. Web-based Boostlets provide a modular architecture and client-side processing capabilities to apply advanced image-processing techniques using consumer-level hardware. The code is open-source and available.
    
} 


\begin{document}



\firstsection{Introduction}
\maketitle
Medical image processing is a common approach to visualizing and comprehending medical data. Different web-based software frameworks exist to visualize the images. However, these frameworks are often limited in processing capabilities and focus mainly on pure visualization tasks. In addition, adding processing algorithms to these frameworks requires targeted software development with a specific programming interface.





To address this, we present Boostlet.js, a tool designed to expand existing web-based visualization frameworks with image processing capabilities (\cref{fig:boost_vs_frameworks}). Boostlet offers a variety of pre-built example functionalities like large language models (LLMs), image filters, and other more advanced algorithms. 
Boostlet also simplifies the development of computational algorithms, making it easy for members of the Medical Imaging Community to create their image processing plugins and share them with others. Boostlet.js integrates seamlessly with popular web-based visualization frameworks such as Cornerstone2D \cite{Ziegler_Open_Health_Imaging}, NiiVue \cite{Niivue}, OpenSeaDragon \cite{OpenSeaDragon}, Xtk \cite{Xtk}, and Papaya \cite{Papaya}. The user only needs to drag and drop the \PB~into the web bookmarks bar. Once clicked on a compatible website, a menu with all available functionalities will be displayed through JavaScript injection.

\begin{table}[h!]
  \centering
    \begin{tabular}{ccc}
     \toprule
 & \textbf{Boostlet} & \textbf{Static Frameworks} \\  \midrule
  \rule{0pt}{1.2\normalbaselineskip}
\textbf{Cross-compatibility?} & \includegraphics[width=10px]{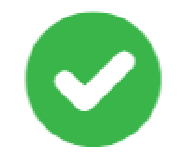} & \includegraphics[width=10px]{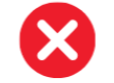} \\ 
\textbf{Easy to share?} & \includegraphics[width=10px]{figures/hook.png} & \includegraphics[width=10px]{figures/cross.png} \\
\textbf{Quick development?} & \includegraphics[width=10px]{figures/hook.png} & \includegraphics[width=10px]{figures/cross.png} \\
\bottomrule
 
\end{tabular}
\\
   \caption{Boostlet allows for easy cross-framework compatibility, sharing, and prototyping with its simple API, while current state-of-the-art frameworks are static, and the development of new functionalities takes time to propagate amongst other popular frameworks.}
    \label{fig:boost_vs_frameworks}
\end{table}

\textbf{Example usage}: a user wants to segment brain-related images on a public data repository like OpenNeuro.org \cite{markiewicz2021openneuro} that visualizes data with NiiVue. Without Boostlet, the user would have to download the data, import it into another compatible tool or model, obtain the segmentation output, and then focus on a single region of interest (ROI). This process would have to be repeated each time the user wants to segment an image. With Boostlet, the user can click on the \PB~bookmark while visiting OpenNeuro.org, select the segmentation tool, draw the region of interest, and immediately view the segmentation overlay without downloading. The exact same process would work on another popular data repository, the Imaging Data Commons from the National Cancer Institute \cite{Schacherer_2023}. 

We built Boostlet.js as a unified programming interface to extend existing web-based visualization libraries with image processing capabilities. In this paper, we describe Boostlet's underlying software architecture and software engineering infrastructure, including automated integration testing and documentation. We then present the \PB~user interface that simplifies the access of existing processing modules and the development of new functionality. Finally, we collect and discuss feedback from experts who develop web-based medical image processing algorithms and test the Boostlet library. All our developments and experiments are available as open science at: \href{https://github.com/mpsych/boostlet}{https://github.com/mpsych/boostlet}.


\section{Related Work}
\label{sec:related_work}

Many frameworks for web-based visualization of biomedical images exist.
Cornerstone2D.js~\cite{Ziegler_Open_Health_Imaging} is acclaimed for its Digital Imaging and Communications in Medicine (DICOM) 
image-rendering abilities and customizability, making it a staple in clinical settings. NiiVue.js~\cite{Niivue} leverages WebGL's power for neuroimaging, providing high-performance interactive visualization through direct access to the graphical processing unit in the browser. OpenSeadragon~\cite{OpenSeaDragon} excels in displaying high-resolution images, essential for visualizing microscope data.
Papaya.js~\cite{Papaya} is a tool dedicated to DICOM and Neuroimaging Informatics Technology Initiative (Nifti) image rendering, providing visualization overlay capabilities and a suite of image control options, similar to the first medical visualization toolkit XTK~\cite{Xtk}. All frameworks offer limited in-built image processing features, a gap now being bridged by Boostlet.js, which extends their functionalities into advanced image processing that can be developed once and used with all of them.


Most related to Boostlets is Brainchop.org \cite{masoud2023brainchop} represents a significant advancement in processing large-scale neuroimaging data using machine learning, offering advanced neural network models applied to brain imaging datasets on the client side. Boostlet.js complements this by providing an accessible framework that facilitates the application of such machine-learning techniques throughout multiple websites. This enables researchers to perform complex analyses with improved efficiency and ease without high-end equipment. Together, Boostlet.js and platforms like Brainchop.org enhance neuroimaging capabilities and ensure compatibility across different frameworks.


\section{Boostlet.js Framework Overview and Examples}

The Boostlet.js library was designed to provide user-friendly installation (as simple as adding a bookmark) for image processing that is independent of a visualization framework.
The library enabled the development of processing modules as plugins. These plugins are called \textit{Boostlets}. 
These plugins can access default functionalities such as pixel data access, filtering, segmenting, and data display with a unified API. This way, developers can wrap their algorithm once and support a variety of visualization toolkits.
We provide a range of Boostlets as examples.

\begin{figure}[h!]
 \centering 
 \includegraphics[width=\columnwidth]{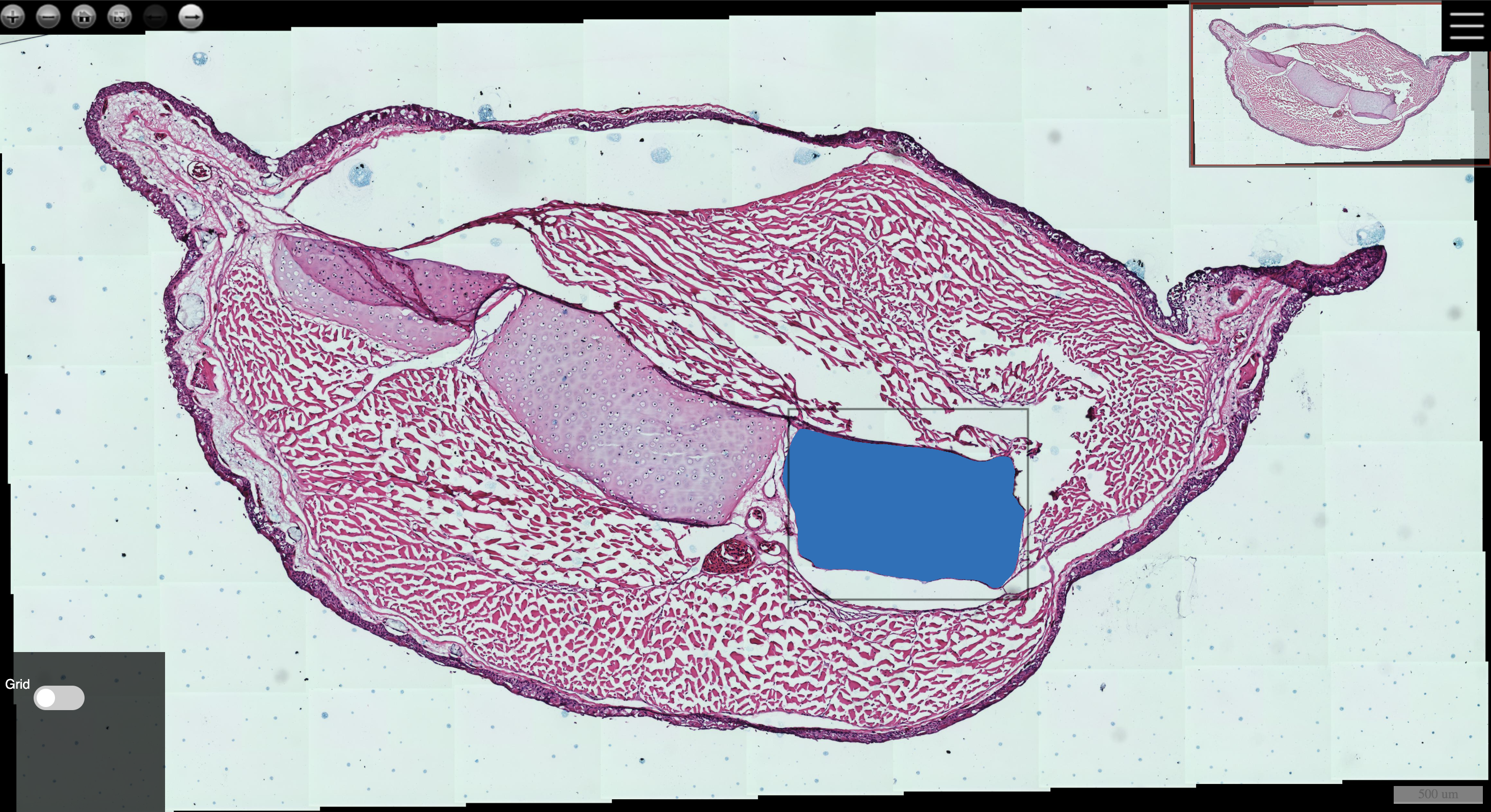}
 \caption{The Segment Anything boostlet executed in \mbox{OpenSeaDragon.js} allows segmenting a region of interest within a slice of microscopy data showing an axolotl limb.}
 \label{fig:boostlets_sam}
\end{figure}

One is Meta's Segment Anything algorithm~\cite{huang2023segment}, which allows the user to segment an area of interest in data displayed by any supported visualization frameworks, as seen in \cref{fig:boostlets_sam}.
For instance, a researcher can search for a relevant case on OpenNeuro.org and segment the corpus callosum by clicking on the Segment Anything Boostlet and dragging a bounding box. The Boostlet will detect NiiVue.js as the underlying visualization framework and query the pixel data. The Boostlet then generates the embedding for the data and computes the segmentation fully on the client side. While Meta provides an ONNX.js compatible predictor module, we also provide the encoding component of Segment Anything in ONNX.js \cite{Microsoft_2018}.
Edge computing with ONNX.js or Tensorflow.js \cite{smilkov2019tensorflowjs} to execute machine learning algorithms allows a wide range of applications. But we also provide examples that use traditional processing and information visualization, such as generating histograms with Plotly.js, Sobel filter-based edge detection~\cite{rorden2022improving}, image captioning powered by Huggingface \cite{huggingface_model}, fiber track decompression using the Trako package~\cite{Trako}, and a melanoma predictor~\cite{kim2024webbased}. Any processing is possible!

\section{User Interface and Experience}


\begin{figure}[h!]
 \centering 
 \includegraphics[width=\columnwidth]{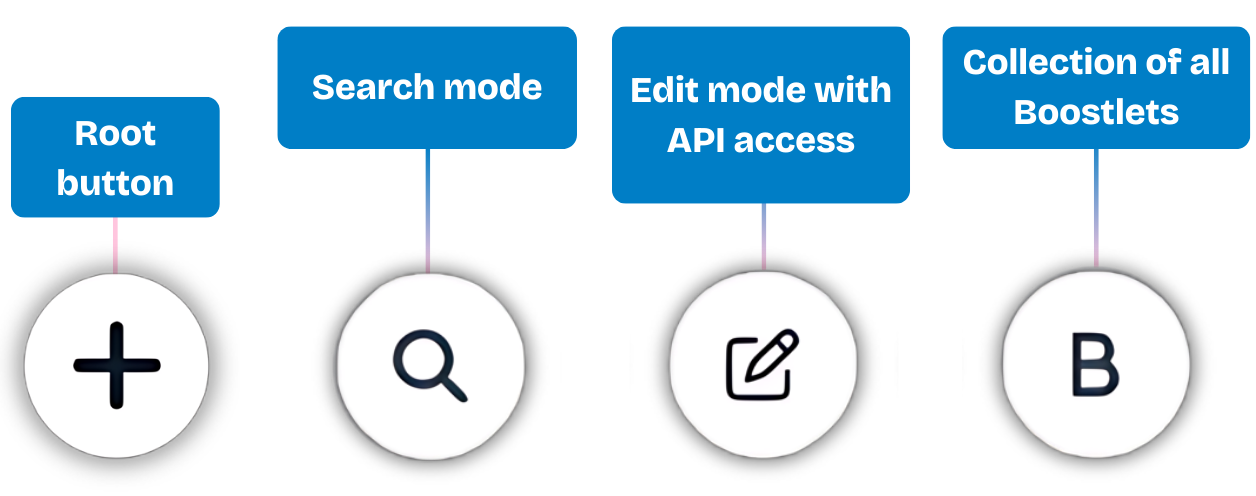}
 \caption{PowerBoost is a versatile plugin that equips any frontend environment with a range of Boostlets and a code editor for API access. Easily activated by dragging and dropping the bookmark from our main website into the bookmarks bar. Upon activation, PowerBoost injects code into the host website, revealing a floating icon. Clicking this icon unveils a user interface that expands into sections for Search, Edit, and Boostlets.}
 \label{fig:powerboostlet}
\end{figure}


The interface features a search bar for quick Boostlet recommendations and a dropdown menu for browsing various categories like Data Visualization, Filters, LLMs, and Machine Learning, with specific examples such as Plotly.js and Sobel. Additionally, the live code editor mode offers a comprehensive panel for executing various functions, including framework detection and HTTP POST requests, enabling users to create and test Boostlets in real time. Thus, accessibility is prioritized.

\section{Development and Integration}

To enhance accessibility for advanced image manipulation, we offer user-friendly functions that integrate various frameworks discussed in \cref{sec:related_work}.

\begin{lstlisting}[language=JavaScript, caption=The Boostlet superclass has a modular design to allow easy extension and integration with various web-based visualization frameworks. 
% TOO WORDY, DUPLICATE Demonstrating Boostlet.js’s adaptability and focus on enhancing web-based image processing capabilities.
, label=lst:boostlet_superclass]
import {Util} from './util.js';
import {Framework} from './framework.js';

export class Boostlet {

  constructor() { }

  init(name, instance) { }

  async select_box(callback) { }

  async select_seed(howmany) { }

  async load_script(url, callback) { }

  async send_http_post(url, data, callback) { }

  get_image(from_canvas) { }

  set_image(new_pixels) { }

  set_mask(new_mask) { }

  convert_to_png(uint8array, width, height) { }

  filter(pixels, width, height, kernel) { }

  hint(message, duration) { }
}
\end{lstlisting}

\cref{lst:boostlet_superclass} shows examples of utility functions from the superclass \textit{Boostlet}. These include \textit{load\_script()} for external script execution, \textit{send\_http\_post()} for HTTP requests to process API calls. There are also data transformation functions such as \textit{convert\_to\_png()}, \textit{grayscale\_to\_rgba()}, \textit{rgba\_to\_grayscale()}, and \textit{harden\_mask()} to enable image manipulation and prepare image data for the final output.

The \textit{Boostlet} superclass also offers functions that allow user interaction during Boostlet execution. For example, \textit{select\_box()} lets the user select a rectangular region of interest (ROI) for manual segmentation or region-based analysis. If a framework does not support a specific mode of interaction (such as OpenSeaDragon.js), the integrated BoxCraft.js~\cite{Varade_2023} library implements custom widgets.
Another interactive function is \textit{set\_mask()}, which applies a mask to the image or canvas, which can be used to highlight, extract, or contrast different types of pixel data. 

\subsection{Framework Integration}
\label{sec:framework_integration}

Boostlets requires a compatible visualization framework to access pixel data and modify the scene or it will use a Canvas fallback.
Boostlet.js then interacts directly with the individual API of the framework, leveraging each framework's native capabilities.
Due to the object-oriented structure of Boostlet, each new framework is treated as a subclass of the superclass Framework displayed in \cref{lst:boostlet_superclass}. 

If the host website does not support a Boostlet-compatible framework, the user can access the \textbf{2D canvas fallback mechanism} and, in the near future, a \textbf{WebGL hook to handle buffers} (\cref{sec:future_work}). This approach ensures that the core functionalities of Boostlets remain operational, providing a consistent user experience across various web environments. CanvasFallback's \textit{get\_canvas()} function dynamically identifies the largest canvas element on the page but can be forced with user selection. The canvas element is the working area for the image processing tasks.


\subsection{Development of Functionalities}
\label{sec:development_functionalities}
To develop functionalities (i.e., segmentation, filtering, etc.), users can utilize the convenience functions to interact with the image data from the host website.

\begin{lstlisting}[language=JavaScript, caption={Sobel filter example that uses some convenience functions to add custom functionality with a kernel for pixel manipulation.}, label=lst:boostlets_sobel]
script = document.createElement("script");
script.type = "text/javascript";
script.src = "https://boostlet.org/dist/boostlet.min.js";

script.onload = run;
document.head.appendChild(script);
eval(script);

function run() {
  
  // detect visualization framework
  Boostlet.init();

  image = Boostlet.get_image();

  kernel = [
    -1, 0, 1,
    -2, 0, 2,
    -1, 0, 1
  ];

  filtered = Boostlet.filter(image.data, image.width, image.height, kernel);

  Boostlet.set_image( filtered );

}
\end{lstlisting}

\cref{lst:boostlets_sobel} shows a use-case scenario for the superclass Boostlet described in \cref{lst:boostlet_superclass} and works as an example of developing image filtering functionalities. This example begins with the initialization process using the \textit{init()} method to detect active visualization frameworks automatically.

After framework detection, Boostlet employs the \textit{get\_image()} function to fetch pixel data from the current image displayed on the canvas. Later, it uses the \textit{filter()} function to execute the pixel transformation with the kernel. Finally, \textit{set\_image()} is used to reflect the pixel manipulation on the canvas. Overall, the core interaction with data relies on \textit{get\_image()} and \textit{set\_image()} functions since these allow users to retrieve the current image displayed by the visualization framework and apply processed data back to the canvas.


\section{Buildsystem and Automated Testing}
For Boostlet.js developers, we use a build system managed with Parcel, a fast, zero-configuration web app bundler~\cite{parcel2018}. The command \verb|npx parcel build| compiles the project into minified Javascript and bundles all assets into a single file suitable for production. In addition, we use node package manager (npm) scripts to manage submodules like the integrated Boxcraft.js~\cite{Varade_2023}, live development server initiation, and testing scripts.

The Boostlet library includes a testing environment for local development, which facilitates collaboration.
\begin{figure}[h!]
 \centering 
 \includegraphics[width=\columnwidth]{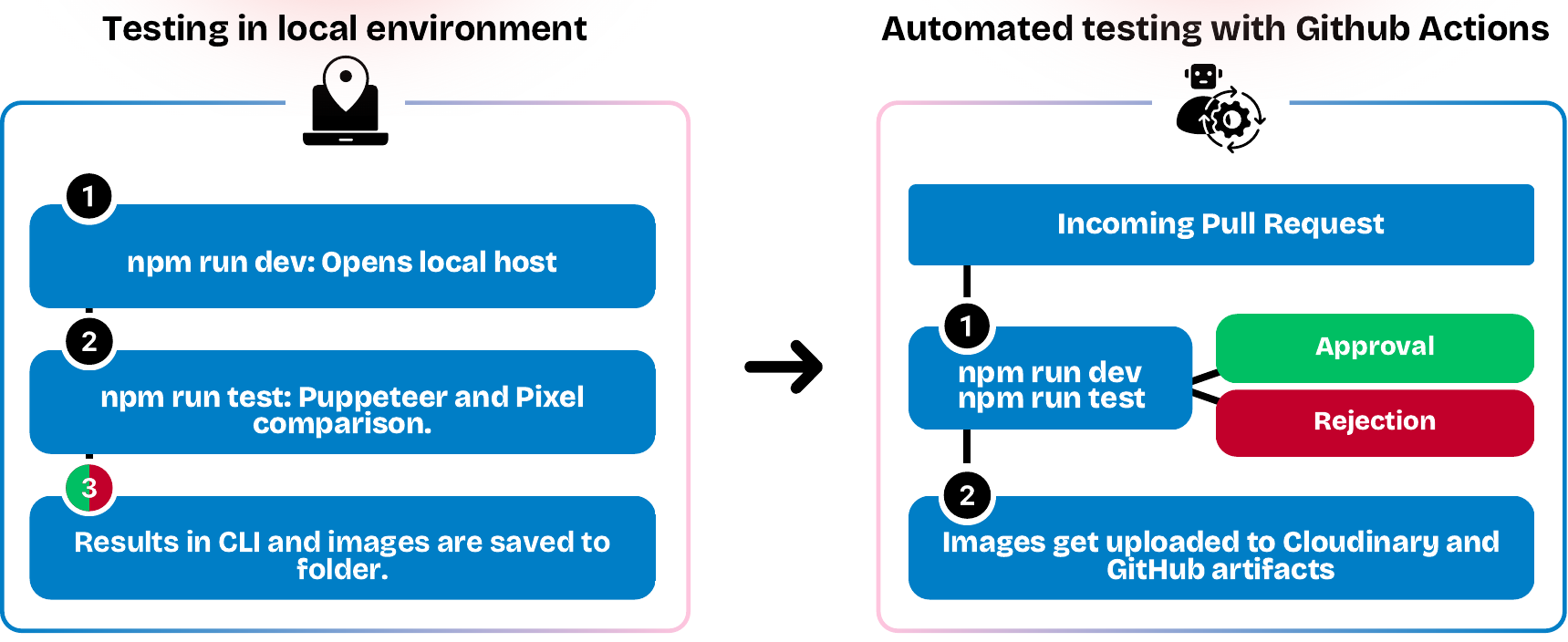}
 \caption{Pipeline for automated local and online testing.}
 \label{fig:testingframework}
\end{figure}
The testing process for Boostlet.js is designed in a multi-step pipeline (\cref{fig:testingframework}) and focuses on validating the integrity of the image processing functionalities within the used framework during the development process.
Testing is performed both locally on the developer's machine and automatically after submitting a pull request on GitHub. The aim is to ensure interoperability and continuous integration of newly added features within the existing codebase.

In the first step, a local server application runs the automated tests, which include the execution of several boostlets and screen captures after each run. In the next step, the produced screenshot is evaluated by Puppeteer~\cite{Puppeteer_2018} to compare pixel-wise against pre-defined ground truth images. Any discrepancy larger than 5\% of the total number of pixels results in a failed test, indicating 
%
%
a possible issue or bug. Any new code is only merged when all tests pass.
%
%


\section{Discussion and Future Work}

\subsection{Community Survey}

\begin{figure}[h!]
 \centering
 \includegraphics[width=\columnwidth]{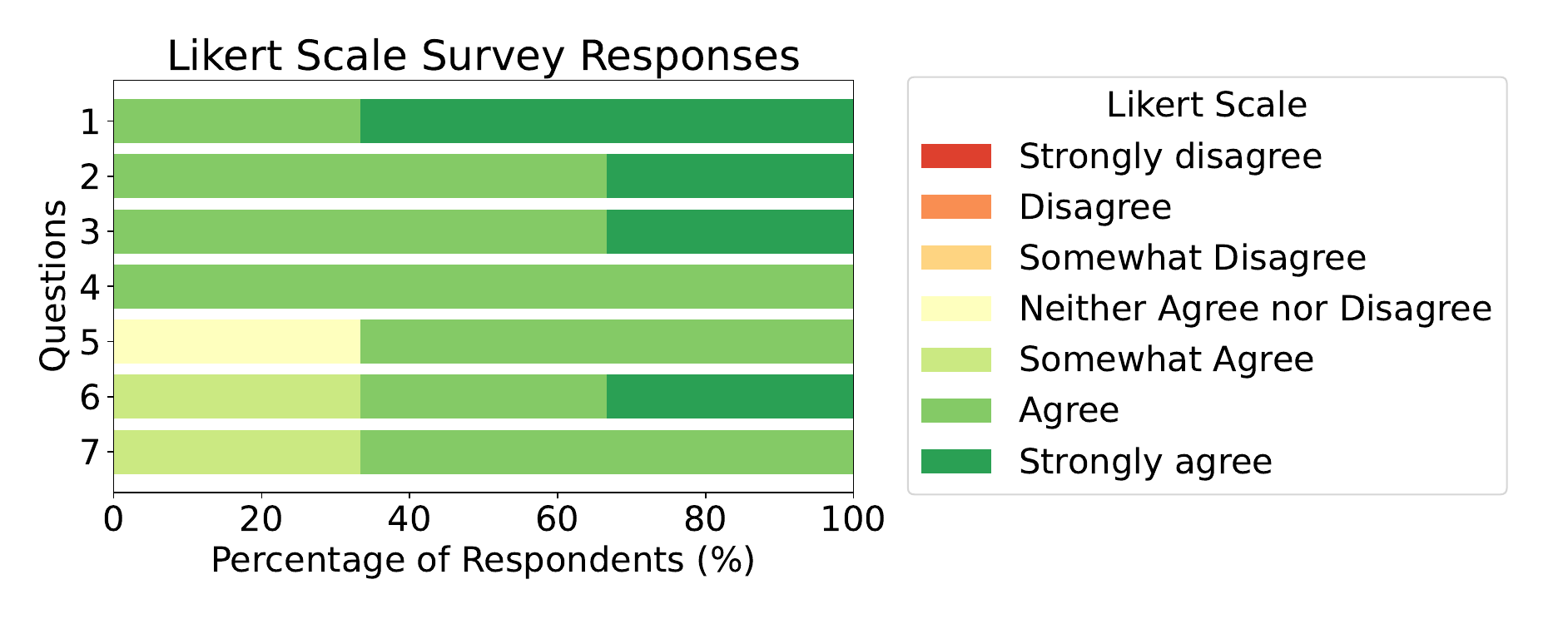}
 \caption{Results of the Likert scale survey show that all users either stay neutral, agree, or strongly agree with the questions prompted in \cref{fig:survey_questions}. *Results for question 8 are not plotted since it is a short-answer feedback question.}
 \label{fig:likertResults_plot}
\end{figure}

We asked a group of medical imaging and visualization professionals of different institutions to provide feedback and evaluation by completing an 8-question survey (\cref{fig:survey_questions}). We received three anonymous responses; the results are in \cref{fig:likertResults_plot}.


\begin{table}[h!]
  \centering
  \begin{tabular}{cp{0.8\linewidth}}
    \toprule
    \textbf{No.} & \textbf{Question} \\
    \midrule
    1 & I found Boostlet.js easy to integrate with existing web-based frameworks. \\
    2 & The process of adding Boostlet.js to my browser as a bookmark is straightforward. \\
    3 & The PowerBoost user interface is easy to navigate. \\
    4 & The client-side processing capabilities of Boostlet.js are satisfactory. \\
    5 & The API for pixel data manipulation provided by Boostlet.js is user-friendly. \\
    6 & Boostlet.js makes it easy to share or develop image processing plugins with the medical imaging community. \\
    7 & I am likely to recommend Boostlet.js to other members of the medical imaging community. \\
    8 & Please provide any additional comments or suggestions for improving Boostlet. \\
    \bottomrule
  \end{tabular}
\\
   \caption{List of questions given to an audience of three medical imaging and visualization professionals regarding their experience with Boostlet.}
   \label{fig:survey_questions}
\end{table}

According to the survey results, Boostlet has a strong user experience and functionality foundation, as seen in responses to questions 1 through 7 (\cref{fig:survey_questions}, \cref{fig:likertResults_plot}). However, for question 8, respondents suggested that Boostlet could improve in making its processes more intuitive, such as allowing the drawing of multiple bounding boxes without refreshing and displaying helpful messages about ongoing processes. Another suggestion was to expand the model library to include a wider range of medical imaging tasks, which is now the current objective. Sharing Boostlet.js with the medical imaging and visualization community could help to achieve this goal.

\subsection{Limitations and Future Work}
\label{sec:future_work}

Boostlet only supports a 2D canvas fallback when the framework is not detected as compatible~\cref{sec:framework_integration}. However, some frameworks use WebGL or WebGPU. Thus, current efforts focus on a WebGL/WebGPU fallback mode that would allow pixel manipulation.
Early experiments confirm that JavaScript injected code can wrap around the WebGL context to monitor all rendering commands sent to the graphical processing unit (GPU). These then can be used to perform pixel data extraction from the framebuffers before data gets rendered. These pixels can then be processed using JavaScript code or in a fragment shader as part of a Boostlet plugin.

\section{Conclusion}
Boostlet.js enhances web-based visualization frameworks with medical image processing capabilities by injecting JavaScript code. This allows to develop a processing plugin (Boostlet) once, that then works with a range of supported visualization frameworks without any further tailoring. Boostlets are installable as bookmarklets or available within the bundled PowerBoost user interface.



\label{sec:supplemental_materials}

\acknowledgments{
The authors wish to thank Jenna Kim, Dhruv Shah, Adnan Ali, Sai Bharadwaj, Likhith Charugundla, Gopi Krishna, Neha Rapolu, Roshan Vemula and our anonymous evaluators for additional testing and improvements. This work was partly supported by
grants from the Alfred P. Sloan Foundation and Oracle for undergraduate research.}

\bibliographystyle{abbrv-doi}

\bibliography{template}
\end{document}